\documentclass[10pt,twocolumn,letterpaper]{article}

\usepackage{wacv}
\usepackage{times}
\usepackage{epsfig}
\usepackage{graphicx}
\usepackage{amsmath}
\usepackage{amssymb}
\usepackage{caption}
\usepackage{dirtytalk}
\usepackage[dvipsnames]{xcolor}
\usepackage{mathrsfs}

\def\wacvPaperID{28} 

\wacvfinalcopy 

\ifwacvfinal
\def\assignedStartPage{1} 
\fi

\ifwacvfinal
\usepackage[breaklinks=true,bookmarks=false]{hyperref}
\else
\usepackage[pagebackref=true,breaklinks=true,colorlinks,bookmarks=false]{hyperref}
\fi

\ifwacvfinal
\else
\fi

\begin{document}

\title{Shadow Art Revisited: A Differentiable Rendering Based Approach}


\author{Kaustubh Sadekar\\ 
        CVIG Lab, IIT Gandhinagar\\ {\tt\small sadelkar.k@iitgn.ac.in}
        \and 
        Ashish Tiwari\\
        CVIG Lab, IIT Gandhinagar\\ {\tt\small ashish.tiwari@iitgn.ac.in}
        \and 
        Shanmuganathan Raman\\
        CVIG Lab, IIT Gandhinagar\\
        {\tt\small shanmuga@iitgn.ac.in}}




\twocolumn[{%
\renewcommand\twocolumn[1][]{#1}%
\maketitle
\begin{center}
    \includegraphics[width=\linewidth, height=4.2cm]{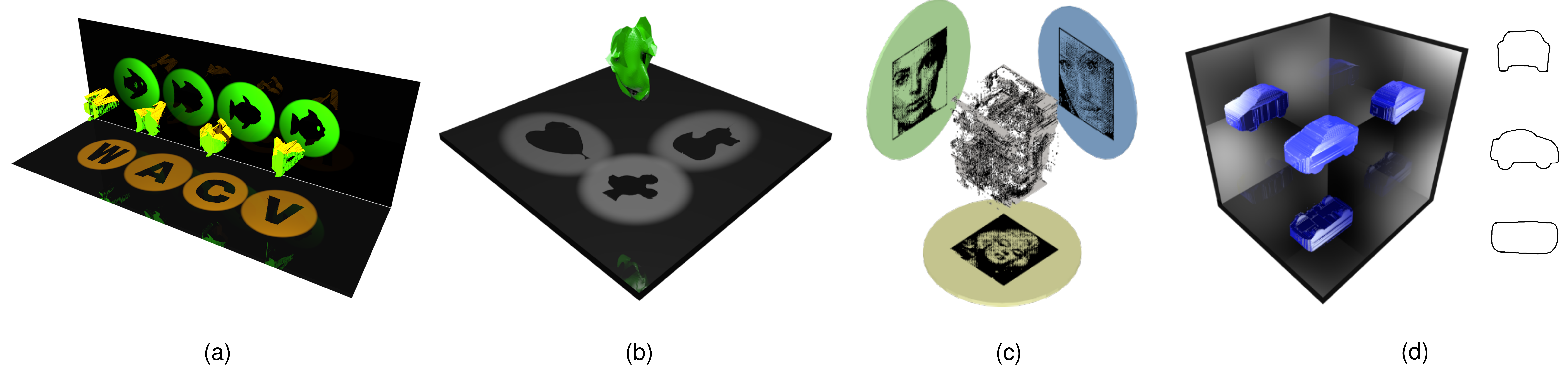}
    \captionof{figure}{Shadow art sculptures generated using differentiable rendering casting the shadows of (a) WACV acronym on one plane and fishes on the other resembling an aquarium of floating objects, (b) dropping \textit{Heart}, \textit{Duck}, and \textit{Mickey} (all on the same plane), and (c) face sketches using half-toned images. (d) 3D reconstruction of a car from hand drawn sketches.}
    \label{fig:teaser}
\end{center}%
}]

\begin{abstract}
   While recent learning based methods have been observed to be superior for several vision-related applications, their potential in generating artistic effects has not been explored much. One such interesting application is Shadow Art - a unique form of sculptural art where 2D shadows cast by a 3D sculpture produce artistic effects. In this work, we revisit shadow art using differentiable rendering based optimization frameworks to obtain the 3D sculpture from a set of shadow (binary) images and their corresponding projection information. Specifically, we discuss shape optimization through voxel as well as mesh-based differentiable renderers. Our choice of using differentiable rendering for generating shadow art sculptures can be attributed to its ability to learn the underlying 3D geometry solely from image data, thus reducing the dependence on 3D ground truth. The qualitative and quantitative results demonstrate the potential of the proposed framework in generating complex 3D sculptures that go beyond those seen in contemporary art pieces using just a set of shadow images as input. Further, we demonstrate the generation of 3D sculptures to cast shadows of faces, animated movie characters, and applicability of the framework to sketch-based 3D reconstruction of underlying shapes.
\end{abstract}

\section{Introduction}
According to an ancient Roman author, Pliny the Elder, \textit{the very art of painting originates from trailing the edge of shadow}. If art can be defined as creating visual or auditory elements that express the author’s imaginative or technical skill, then shadow art represents those skills in play with shadows. Most of us have seen or atleast heard of “someone” making “something” interesting out of shadows. However, it is usually limited to people playing around with their fingers around a lamp, making shadows of rabbits or horses on the wall. In this work, we show how differentiable rendering can  be used to generate some amazing 3D sculptures which cast mind-boggling shadows when lit from different directions. 

\begin{figure}[h]
    \centering
    \includegraphics[width=\linewidth]{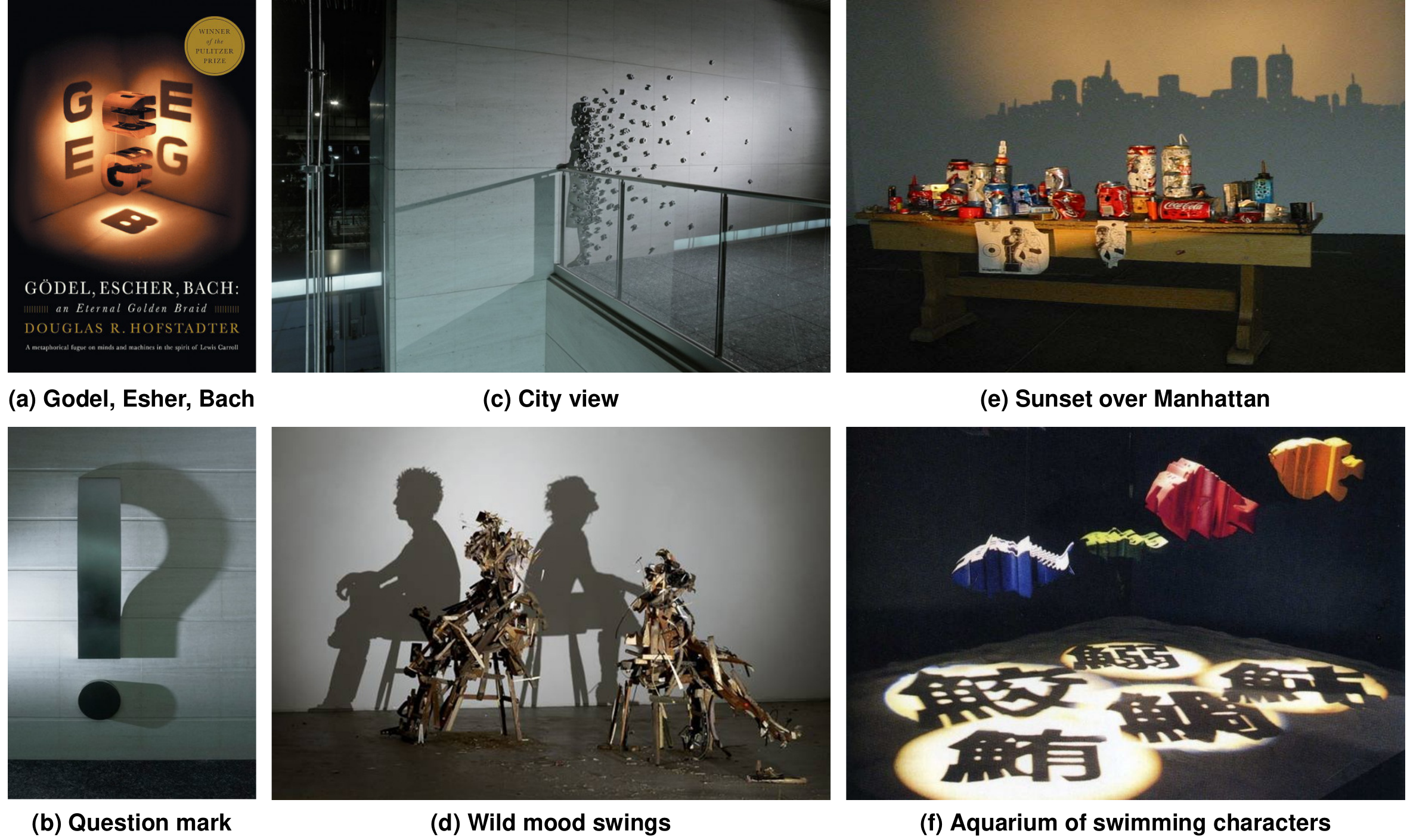}
    \caption{Examples of shadow art sculptures by (a) \textit{Douglas Hofstadter}, (b, c) \textit{Kumi Yamashita}, (d, e) \textit{Tim Noble} and \textit{Sue Webster}, and (f) \textit{Shigeo Fukuda}.}
    \label{fig:examples}
\end{figure}

Figure \ref{fig:examples} (a) shows the cover of the book \textit{Gödel, Escher, Bach} by \textit{ Douglas Hofstadter} that features blocks casting shadows of different letters when seen from different sides. \textit{Kumi Yamashita} - one of the most prominent contemporary artists - demonstrated that seemingly simple objects arranged in a certain pattern cast startling silhouettes when lit from just the right direction. An exclamation point becomes a question mark when lit from its side (Figure \ref{fig:examples} (b)) and a bunch of aluminum numbers placed in a 30-story office add up to an image of a girl overlooking the crowd below (Figure \ref{fig:examples} (c)). All of these, and other pieces made by \textit{Kumi Yamashita} not only please our eyes, but also inspire emotion and pose intriguing questions. \textit{Tim Noble} and \textit{Sue Webster} have been presenting this type of artwork since 1997, creating projected shadows of people in various positions (Figure \ref{fig:examples} (d)). This specifically arranged ensemble shows how readily available objects can cast the clearest of illusions of clearly recognizable scenes (Figure \ref{fig:examples}(e)). Figure \ref{fig:examples} (f) shows the aquarium of floating characters by \textit{Shigeo Fukuda} where the shadows of the fish reveal their names in kanji characters. Even after such fascinating effects, the current state of shadow art seems to be well described by Louisa May Alcott, who says \textit{“Some people seemed to get all sunshine, and some all shadow…”}. Shadow art was first introduced to the vision and graphics community by \cite{mitra2009shadow} where they formally addressed the problem in an optimization framework. Since then, no significant progress has been observed in this direction. 

\textbf{The question:} Can we develop a method that learns to create or optimize to 3D sculptures that can generate such artistic effects through their shadows? In this work, we attempt to answer this question through the use of \textit{Differentiable Rendering}. Here, instead of trying to realistically render a scene of our creation, we try to reconstruct a representation of a scene from one or more images of a scene \cite{kato2020differentiable}. Our work is mostly inspired by examples of shadow art shown in Figure \ref{fig:examples}. Specifically, our objective is to generate 3D shadow art sculptures that cast different shadows (of some recognizable objects) when lit from different directions using a differentiable renderer.

\textbf{Why differentiable rendering?} Most learning-based methods for 3D reconstruction require accurate 3D ground truths as supervision for training. However, all we have is a set of desired shadow images in our case. Differentiable rendering based methods require only 2D supervision in the form of single or multi-view images to estimate the underlying 3D shape, thus, eliminating the need for any 3D data collection and annotation.

\textbf{Contributions.}
The following are the major contributions of this work.
\begin{itemize}
    \item We introduce the creation for 3D shadow art sculptures that cast different shadows when lit from different directions using differentiable rendering just from the input shadow images and the corresponding projection information. 
    \item We demonstrate the efficacy of deploying differentiable rendering pipeline over voxel and mesh based representations to generate shadow art sculptures.
    \item We show that the proposed framework can create artistic effects that go beyond those seen in contemporary art pieces by generating 3D sculptures using half-toned face images and its sketches drawn from multiple viewpoints.
    \item To the best of our knowledge, ours is the first work to address shadow art using differentiable rendering.
\end{itemize}

\textbf{Organization.} We start by describing the literature covering the relevant related works in Section \ref{sec:related_work}. We discuss the problem statement more formally and describe both voxel and mesh based differentiable rendering pipelines in Section \ref{sec:method}. Section \ref{sec:loss_function} describes the loss functions deployed for optimization. In Section \ref{sec:experiments}, we perform qualitative and quantitative analysis of results and compare the performance of the proposed framework with that of the state-of-the-art method. Section \ref{sec:applications} describes interesting artistic effects and applications of shadow art before we conclude the work in Section \ref{sec:conclusion}.

\section{Related Work}
\label{sec:related_work}
Shadows play an essential role in the way we perceive the world
and have been central in capturing the imagination of many artists including stage performers. Several artists have typically used manual, trial-and-error style approaches to create 3D shadow sculptures. However, with the advent of digital design technology, the need of an automated framework is inevitable.

\textbf{Shadow Art.} Shadows in many computer graphics and computer vision applications have been studied from both perceptual (artist's) and mathematical (programmer's) point of views. It started with studying the effect of shadow quality on perception of spatial relationships in a computer generated image \cite{wanger1992effect, wanger1992perceiving}. Pellacini \emph{et al.} developed an interface for interactive cinematic shadow design that allows the user to modify the positions of light sources and shadow blockers by specifying constraints on the desired shadows \cite{pellacini2002user}. The idea of computing the shadow volume from a set of shadow images evolved after that. This is similar to the construction of a visual hull used for 3D reconstruction. Visual hull is the intersection volume of a set of generalized cones constructed from silhouette images and the corresponding camera locations \cite{laurentini1994visual}. Sinha and Polleyfeys \cite{sinha2005multi} studied the reconstruction of closed continuous surfaces from multiple calibrated images using min-cuts with strict silhouette constraints.

\textbf{Relation with the state-of-the-art method.} The work closest to ours is by Mitra \emph{et al.} \cite{mitra2009shadow}. They described shadow art more formally by introducing a voxel-based optimization framework to recover the 3D shape from arbitrary input images by deforming the input shadow images and handled inherent image inconsistencies. In this work, we demonstrate the potential of differentiable rendering in generating 3D shadow sculptures all from the arbitrary shadow images without any explicit input image deformation. Although the associated 3D object might not exist in the real-world, but the method still creates shadow sculptures that go beyond those seen in contemporary art pieces casting the physically realizable shadows when lit from appropriate directions.

\textbf{Differentiable Rendering.} We briefly review methods that learn the 3D geometry via differentiable rendering. These methods are categorized based on the underlying representation of 3D geometry: point clouds, voxels, meshes, or neural implicit representation. In this work, we primarily focus on voxel and mesh based representations. 

Several methods operate on voxel grids \cite{lombardi2019neural, nguyen2018rendernet, paschalidou2019superquadrics, tulsiani2017multi}. Paschalidou \emph{et al.} \cite{paschalidou2019superquadrics} and Tulsiani \emph{et al.} \cite{tulsiani2017multi} propose a probabilistic ray potential formulation. Although they provide a solid mathematical framework, all intermediate evaluations need to be saved for backpropagation. This limits these approaches to relatively small-resolution voxel grids. On one hand, Sitzmann \emph{et al.} \cite{sitzmann2019scene} have inferred implicit scene representations from RGB images via an LSTM-based differentiable renderer and Liu \emph{et al.} \cite{liu2019learning} perform max-pooling over the intersections of rays with a sparse number of supporting regions from multi-view silhouettes. On the other hand, \cite{niemeyer2020differentiable} show that volumetric rendering is inherently differentiable for implicit representations and hence, no intermediate results need to be saved for the backward pass. OpenDR \cite{loper2014opendr} roughly approximates the backward pass of the traditional mesh-based graphics pipeline. Liu \emph{et al.} \cite{liu2019soft} proposed Soft Rasterizer to replace the rasterization step with a soft version to make it differentiable using a deformable template mesh for training and yields compelling results in reconstruction tasks. We deploy this in our mesh-based differentiable rendering pipeline for rasterization.

Both voxel and mesh based representations have their own strength and weaknesses. In this work, we describe the differentiable rendering optimization framework for both these 3D representation and discuss which model fits the best for different scenarios to create plausible shadow art sculptures


\begin{figure*}[ht]
    \centering
    \includegraphics[width=0.9\linewidth]{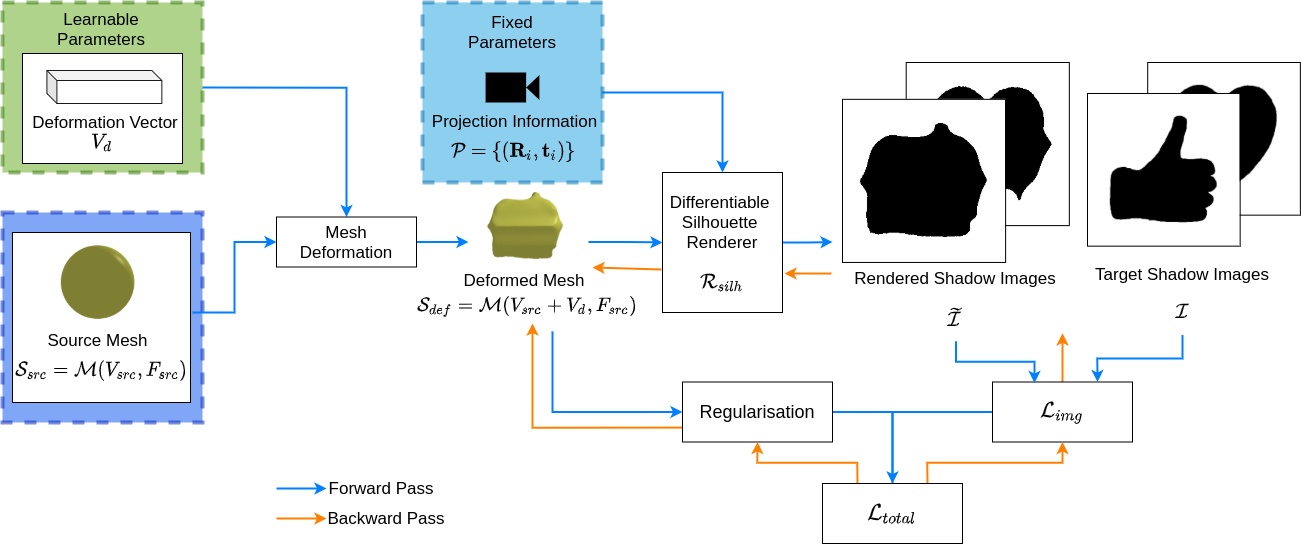}
    \caption{Information flow in the proposed mesh-based differentiable rendering pipeline.}
    \label{fig:block_diagram}
\end{figure*}
\section{Method}
\label{sec:method}

\subsection{Problem Formulation}
\label{prob_formulation}
The key idea of our work is to generate an artistic 3D sculpture $\mathcal{S}$ that casts $N$ different shadows when lit from $N$ different directions using differentiable rendering based optimization pipeline. The prime focus here is to create interesting shadow art effects using the 3D sculpture $\mathcal{S}$. The input to the pipeline is a set $\mathcal{X} = \{X_{1}, X_{2}, ..., X_{N}\}$ of shadow configuration $X_{i} = (I_{i}, P_{i})$. $I_{i}$ represents the target shadow image and $P_{i}$ is the corresponding projection information.  

The shadow of an object can be regarded as its projection on a planar surface. Assuming directional lighting, this projection is an \textit{orthographic projection} when the surface is perpendicular to the lighting direction and a \textit{perspective projection}, otherwise \cite{Abbott1971}. Obtaining shadow of an object is equivalent to finding the corresponding silhouette captured by a camera pointing in the same direction as the light source. Therefore, $I_{i}$ the shadow image, is essentially a silhouette. From here on, we shall use the term silhouette images and shadow images, interchangeably.

The shadow art problem is similar to a multi-view 3D reconstruction problem \cite{lee2003silhouette, mulayim2003silhouette}, where we try to estimate the 3D structure of an object given its $N$ silhouette views. However, the key differences in shadow art are: (i) the $N$ views can correspond to arbitrary silhouettes (not necessarily of the same object) and (ii) the learned 3D sculpture may bear no resemblance with any real-world object and just be an abstract art that casts the desired shadows when lit from appropriate directions. Undoubtedly, there exist multiple 3D shapes that can cast the same set of shadows. However, our concern is just to learn one such 3D sculpture that can create the desired artistic effects through its shadows.

\subsection{System Overview}

By providing shadow configuration $\mathcal{X} = \{X_{i} = (I_{i}, P_{i})| i=1,2,..., N\}$ as input to the pipeline, the objective is to learn the underlying 3D sculpture $\mathcal{S}$, as described earlier. The projection information $P_{i}$ corresponds to the camera position (and hence, the light source position) associated with $i^{th}$ shadow image $I_{i}$ such that $P_{i} = (\mathbf{R}_{i}, \mathbf{t}_{i})$. Here, $\mathbf{R}_{i}$ and $\mathbf{t}_{i}$ are the 3D rotation and translation of the camera, respectively. We start by initialising $\mathcal{S}$ with a standard geometry which is further optimized by minimizing image-based losses, such that the rendered silhouette images $\widetilde{I}_{i} = I_{i}$ for all $i = 1, 2, ..., N$. The prime reason for using differentiable rendering is that it allows gradient flow  directly from images back to parameters of $\mathcal{S}$ to optimize it in an iterative fashion. In other words, it does not require any explicit 3D supervision and optimizes the 3D shape solely from image based losses. For further simplicity, let the set of target shadow images and the associated projection information be denoted as $\mathcal{I}$ and $\mathcal{P}$, respectively, such that $\mathcal{I} = \{I_{1}, I_{2},..., I_{N}\}$ and $\mathcal{P} = \{P_{1}, P_{2},..., P_{N}\}$. Further, let $\widetilde{\mathcal{I}} = \{\widetilde{I}_{1}, \widetilde{I}_{2},..., \widetilde{I}_{N}\}$ be the set of shadow images obtained from learned 3D sculpture $\mathcal{S}$ as per projections $\mathcal{P}$.

In this work, we consider two common representations for 3D shapes i.e. \textit{voxel} and \textit{mesh} based representations. In the following section, we elaborate the optimization pipelines for voxel and mesh based representations of the 3D object to create visually plausible shadow art using differentiable rendering.

\subsection{Voxel Based Optimization}

In this section, we look at a differentiable rendering pipeline that uses voxels to represent the 3D geometry. A voxel is a unit cube representation of a 3D space. The 3D space is quantized to a grid of such unit cubes. It is parameterized by an $N$-dimensional vector containing information about the volume occupied in 3D space. Additionally, it encodes occupancy, transparency, color, and material information. Even though occupancy and transparency probabilities (in the range $[0,1]$) are different, they can be interpreted in the same way in order to maintain differentiability during the ray marching operation \cite{kato2020differentiable}. A typical rendering process involves collecting and aggregating the voxels located along a ray and assigning a specific color to each pixel based on the transparency or the density value. All the voxels that are located along a ray projecting to a pixel are taken into account when rendering that pixel. However, our objective is to do the inverse, i.e., to find the 3D geometry associated with silhouettes corresponding to different directions. 

We assume that the 3D object $\mathcal{S}$ is enclosed in a 3D cube of known size centered at the origin. Hence, $\mathcal{S}$ can be defined by a learnable 3D tensor $V$ that stores the density values for each voxel. We initialize $V$ with all ones. The color value for each voxel is set to 1 and is kept fixed in the form of a color tensor $C$. Next, we render $\mathcal{S}$ using a differentiable volumetric rendering method described in \cite{ravi2020pytorch3d}. To restrict the voxel density values to the range $[0,1]$, $V$ is passed through a sigmoid activation function $(\sigma)$  to obtain $\widetilde{V}$, as described in Equation \ref{eq:sigmoid}. 

\begin{equation}\label{eq:sigmoid}
\centering
    \widetilde{V} = \sigma(V)
\end{equation}

We then pass $\widetilde{V}$ through the differentiable volume renderer $\mathcal{R}_{vol}$ along with the fixed color tensor $C$ and the associated projection information $\mathcal{P}$ to obtain the set of corresponding rendered images $\widetilde{I}$, as described in Equation \ref{eq:voxrender}.

\begin{equation} \label{eq:voxrender}
\centering
     \widetilde{\mathcal{I}} = \mathcal{R}_{vol}(\widetilde{V}, C, \mathcal{P})
\end{equation}

The voxel densities $V$ are optimized by minimizing the image level loss between a set of rendered shadow images $\widetilde{\mathcal{I}}$ and the corresponding target shadows in $\mathcal{I}$. The image level loss $\mathcal{L}_{img}$ is a weighted combination of $L_{1}$ and $L_{2}$ losses, as described in Equation \ref{eq:imgloss}. 

\begin{equation}\label{eq:imgloss}
    \mathcal{L}_{img} = \lambda_{1}\mathcal{L}_{L_{1}} + \lambda_{2}\mathcal{L}_{L_{2}}
\end{equation}
Here, $\lambda_{1} = 10.0$ and $\lambda_{2} = 10.0$ are the weights associated with $L_{1}$ and $L_{2}$ losses, respectively. The resulting voxel based representation of $\mathcal{S}$ can finally be converted to a 3D mesh making it suitable for 3D printing. One simple way to achieve this is by creating faces around each voxel having density greater than a certain threshold value (as described in \cite{ravi2020pytorch3d}).

\subsection{Mesh Based Optimization}

In this section, we also propose to use mesh based differentiable rendering to meet our objective. The entire workflow is described in Figure \ref{fig:block_diagram}. The 3D object $\mathcal{S}$ can be represented as a mesh $\mathcal{M}(V, F)$. Here, $V$ is a set of vertices connected by a set of triangular faces $F$ that define the surface of $\mathcal{S}$.

We start by initializing a source mesh $\mathcal{S}_{src}$ = $\mathcal{M}(V_{src}, F_{src})$ with an icosphere consisting of $|V_{src}|$ vertices and $|F_{src}|$ faces. The idea is to learn the per-vertex displacements $V_{d}$ to deform $\mathcal{S}_{src}$ to the final desired mesh that casts desired shadows (silhouettes), when lit from appropriate directions. This is achieved by rendering the deformed mesh $\mathcal{S}_{def} = \mathcal{M}(V_{def}, F_{def})$ through a mesh-based differentiable silhouette renderer $\mathcal{R}_{silh}$ (as described in \cite{ravi2020pytorch3d}) from the associated projection $\mathcal{P}$ such that,
\begin{equation}
    \centering
    \begin{split}
    & V_{def} = V_{src} + V_{d} \\ 
    & F_{def} = F_{src} \\
    & \widetilde{\mathcal{I}} = \mathcal{R}_{silh}(\mathcal{S}_{def}, \mathcal{P})
    \end{split}
    \label{eq:silhrender}
\end{equation}

\subsubsection{Loss Function}\label{sec:loss_function}
The source mesh is optimized by minimizing image level loss $\mathcal{L}_{img}$ (described in Equation \ref{eq:imgloss}), normal consistency loss, and imposing Laplacian and edge length regularisation.\\
\\
\textit{\textbf{Normal consistency.}} We use normal consistency loss to ensure smoothness in the resulting 3D sculpture. For a mesh $\mathcal{M}(V,F)$, let $e = (\mathbf{v}_{x}, \mathbf{v}_{y})$ be the connecting edge of two neighboring faces $f_{x} = (\mathbf{v}_{x}, \mathbf{v}_{y}, \mathbf{a})$ and $f_{y} = (\mathbf{v}_{x}, \mathbf{v}_{y}, \mathbf{b})$, such that $f_{x}, f_{y} \in F$ with normal vectors $\mathbf{n}_{x}$ and $\mathbf{n}_{y}$, respectively. If $\widetilde{\mathcal{E}}$ is the set of all such connecting edges $e$ and $|F|$ is the total number of faces in mesh, the normal consistency over all such neighbouring faces $f_{x}$ and $f_{y}$ is given as per Equation \ref{eq:normal_cons}.

\begin{equation}\label{eq:normal_cons}
    \centering
    \mathcal{L}_{norm} = \frac{1}{|F|}\sum_{e \in \widetilde{\mathcal{E}}}(1 - cos(\mathbf{n}_{x}, \mathbf{n}_{y}))
\end{equation}
where,
\begin{equation*}
    \centering
    \begin{split}
    & \mathbf{n}_{x} = (\mathbf{v}_{y} - \mathbf{v}_{x}) \times (\mathbf{a} - \mathbf{v}_{x})\\
    & \mathbf{n}_{y} = (\mathbf{b} - \mathbf{v}_{x}) \times (\mathbf{v}_{y} - \mathbf{v}_{x}).
    \end{split}
\end{equation*}
\\
\textit{\textbf{Laplacian regularisation.}} In order to prevent the model from generating large deformations, we impose uniform Laplacian smoothing \cite{nealen2006laplacian}, as described by Equation \ref{eq:laplacian}.

\begin{equation}\label{eq:laplacian}
    \centering
    \mathcal{L}_{lap} = \frac{1}{|V|}\sum_{i=1}^{|V|}\left(\bigg\lVert\sum_{\mathbf{v}_{j} \in \mathcal{N}(\mathbf{v}_{i})}w_{ij}\mathbf{v}_{j} - \mathbf{v}_{i}\bigg\rVert_{1}\right)
\end{equation}
Here, $|V|$ is the number of vertices in the mesh $\mathcal{M}$ and $\mathcal{N}(\mathbf{v}_{i})$ is the neighbourhood of vertex $\mathbf{v}_{i}$. 
\begin{equation*}
    \centering
    w_{ij} = \frac{\omega_{ij}}{\sum_{k \in \mathcal{N}(i)} \omega_{ik}}
\end{equation*}
For uniform Laplacian smoothing, $\omega_{ij} = 1$, if $(\mathbf{v}_{i}, \mathbf{v}_{j})$ form an edge, $\omega_{ij} = -1$ if $i = j$, and $\omega_{ij} = 0$, otherwise.\\
\\
\textit{\textbf{Edge length regularisation.}} Edge-length regularisation is included to prevent the model from generating flying vertices and is given by Equation \ref{eq:edge_len}.
\begin{equation}\label{eq:edge_len}
    \centering
    \mathcal{L}_{edge} = \sum_{i=1}^{|V|}\sum_{\mathbf{v}_{j} \in \mathcal{N}(\mathbf{v}_{i})}\parallel\mathbf{v}_{i} - \mathbf{v}_{j}\parallel_{2}^{2}
\end{equation}\\
Finally, the overall loss function is as described in Equation \ref{eq:overall_loss}.
\begin{equation}\label{eq:overall_loss}
    \centering
    \mathcal{L}_{total} = \lambda_{a}\mathcal{L}_{img} + \lambda_{b}\mathcal{L}_{norm} + \lambda_{c}\mathcal{L}_{lap} + \lambda_{c}\mathcal{L}_{edge}
\end{equation}
Here, $\lambda_{a} = 1.6$, $\lambda_{b} = 2.1$, $\lambda_{c} = 0.9$, and $\lambda_{d} = 1.8$ are the weights associated with the losses $\mathcal{L}_{img}$, $\mathcal{L}_{norm}$, $\mathcal{L}_{lap}$, and $\mathcal{L}_{edge}$, respectively.

\subsection{Implementation Details} 
The aforementioned differentiable rendering pipelines are implemented using Pytorch3D \cite{ravi2020pytorch3d}. As an initialisation for mesh, we use a level 4 icosphere composed of 2,562 vertices and 5,120 faces. For the voxel based rendering pipeline, we assume that the object is inside a cube (a grid of $128 \times 128 \times 128$ voxels) centered at origin with side of length 1.7 world units. We train the optimization pipeline with custom silhouette images of size $128 \times 128$ for 2000 epochs.  We keep the learning rate to $1 \times 10^{-4}$. We keep the learning rate to $1 \times 10^{-2}$ and train the optimization pipeline for 500 epochs. The training is performed on NVIDIA Quadro RTX 5000 with 16 GB memory.

\section{Experimental Analysis}\label{sec:experiments}
In this section, we perform an extensive analysis over the results obtained using voxel and mesh based differentiable rendering pipelines to create plausible shadow art effects. We start by discussing the evaluation metrics and perform ablation studies to understand the effect of various loss terms in the design.

\subsection{Evaluation Metrics}
Following our discussion in Section \ref{prob_formulation}, we assess the quality of silhouettes (shadow images) obtained through the 3D sculpture $\mathcal{S}$ as per projections $\mathcal{P}$. To compare the rendered silhouette images with the target silhouette images (representing shadows), we use Intersection over Union (IoU) and Dice Score (DS). Additionally, we need to quantify the quality of the 3D sculpture $\mathcal{S}$ obtained after optimization. While we do not have any ground truth for 3D shape, and this is an optimization framework, we need a "no reference" quality metric. Therefore, we decided to use normal consistency evaluated over $\mathcal{S}$ to assess the quality of the mesh.
\begin{figure*}[ht]
    \centering
    \includegraphics[width=\linewidth]{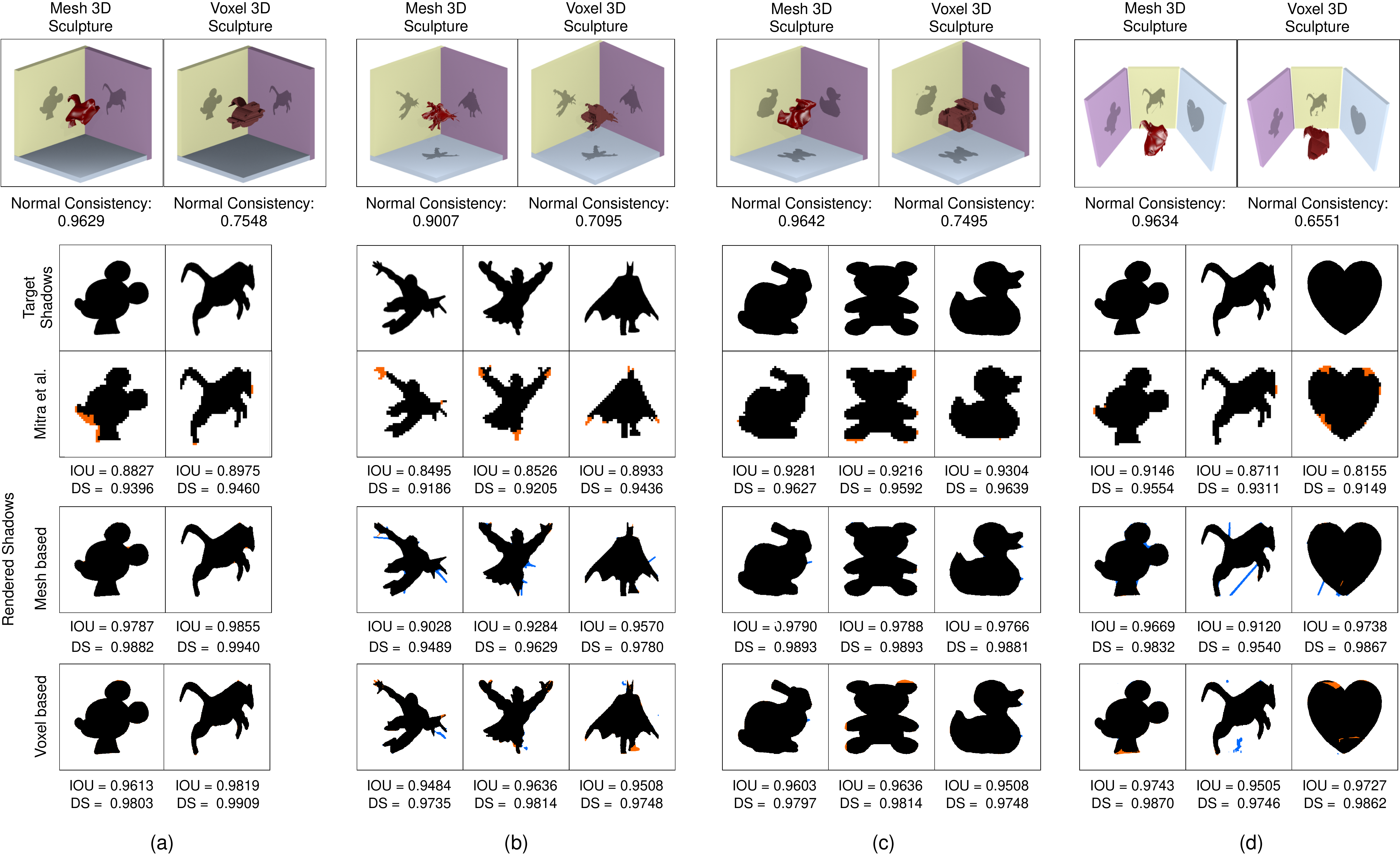}
    \caption{Qualitative and quantitative results on (a) two views (b,c) three orthogonal views, and (d, e) three non-orthogonal views using voxel and mesh-based rendering for shadow art.}
    \label{fig:res}
\end{figure*}

\subsection{Ablation Study}
\begin{figure}[h]
    \centering
    \includegraphics[width=\linewidth, height=2.5cm]{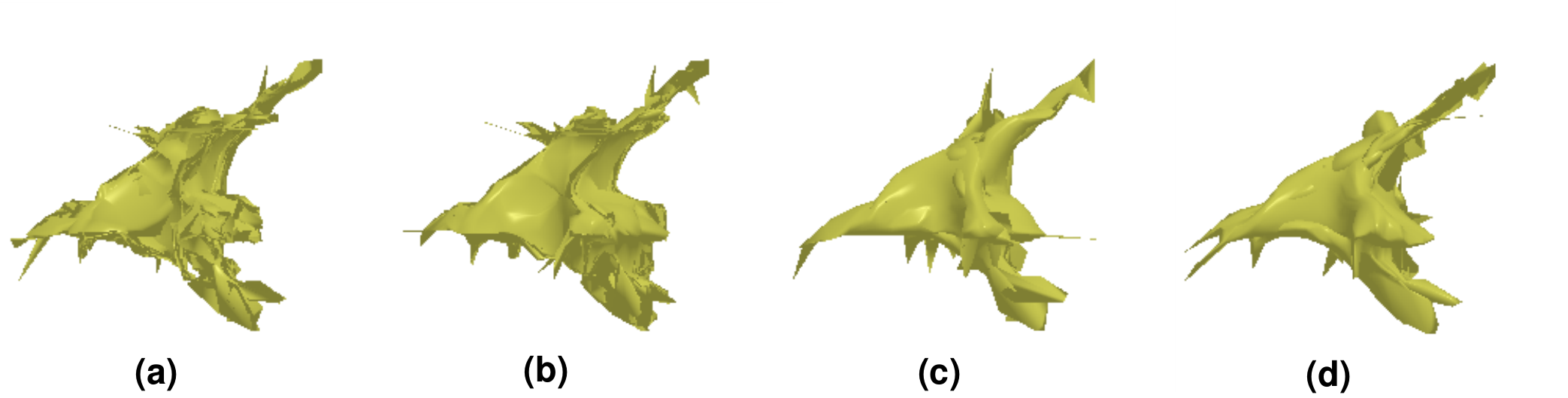}
    \caption{Qualitative analysis of effect of various loss terms. (a) $\mathcal{L}_{img}$, (b) $\mathcal{L}_{img} + \mathcal{L}_{edge}$, (c) $\mathcal{L}_{img} + \mathcal{L}_{edge} + \mathcal{L}_{lap}$, and (d) $\mathcal{L}_{img} + \mathcal{L}_{edge} + \mathcal{L}_{lap} + \mathcal{L}_{norm}$. }
    \label{fig:ablation}
\end{figure}

Figure \ref{fig:ablation} depicts the qualitative effect of different loss terms used in the optimization pipeline. The underlying mesh in this figure corresponds to the arrangement shown in Figure \ref{fig:res} (c). The image based loss $\mathcal{L}_{img}$ alone is not sufficient for generating plausible 3D sculptures as they are expected to suffer from distortions due to flying vertices (spike-like structures in Figure \ref{fig:ablation} (a)) or large deformations. Since we do not have any ground truth for explicit 3D supervision, we examine the effect of including regularisation in the objective function. Figure \ref{fig:ablation} (b) shows that the spikes are reduced by introducing edge-length regularisation. Further, as shown in Figure \ref{fig:ablation} (c), Laplacian smoothing prevents the sculpture from experiencing super large deformations. Finally, normal consistency loss ensures further smoothness in the optimized surface. Figure \ref{fig:ablation} (d) shows the result obtained by applying all the aforementioned regularization terms along with the image based loss. The resulting quality of the mesh validates our choice of loss terms.
\begin{figure*}[ht]
    \centering
    \includegraphics[width=\linewidth]{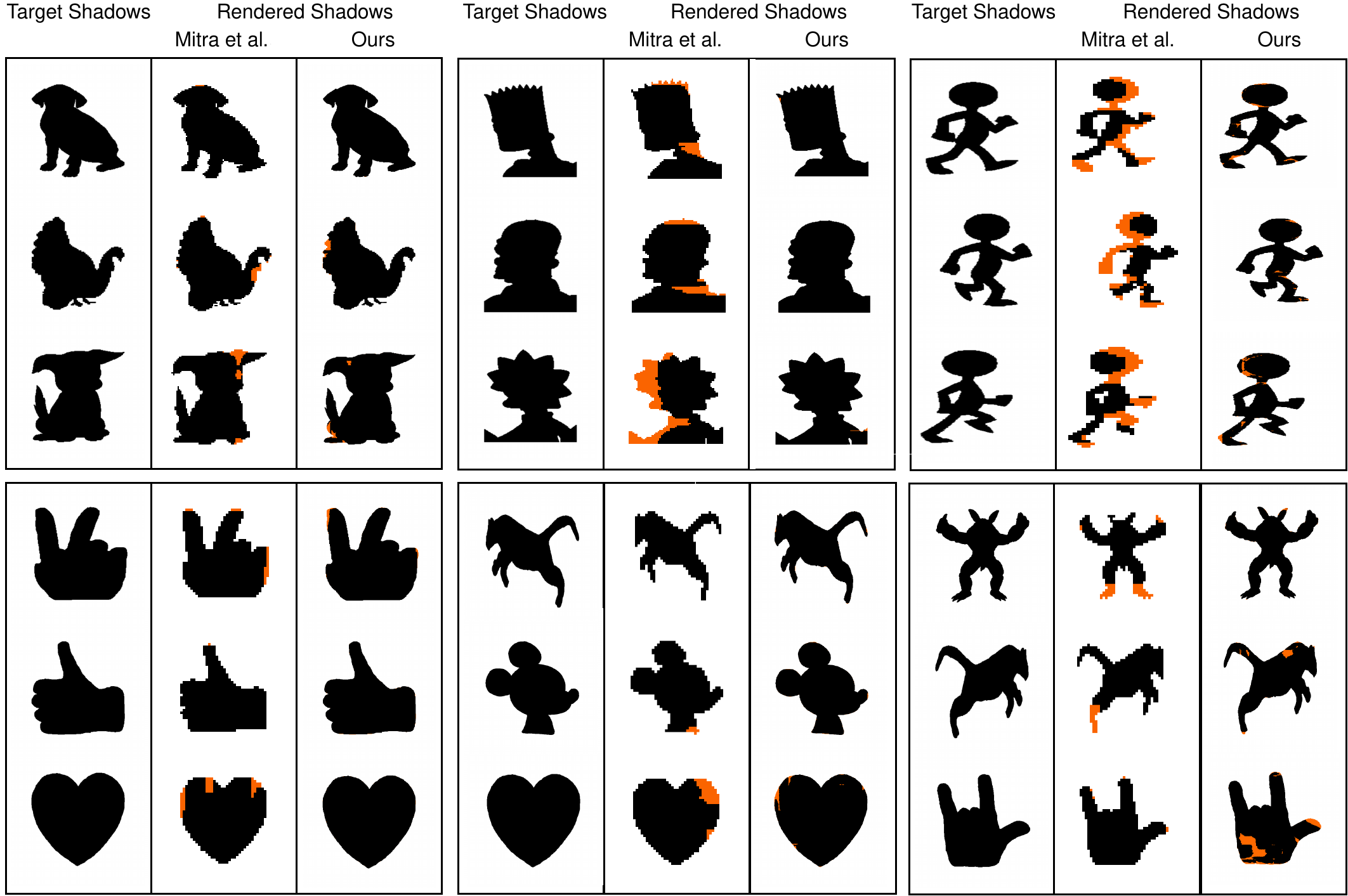}
    \caption{Qualitative evaluation of results obtained through (A) shadow art tool in \cite{mitra2009shadow} and (B) our voxel based rendering pipeline. The inconsistencies are highlighted in orange color.}
    \label{fig:comparison}
\end{figure*}

\subsection{Qualitative and Quantitative Analysis}
In this section, we perform the qualitative and quantitative evaluation on a wide variety of shadow images including those used in \cite{mitra2009shadow} to illustrate the versatility of our approach in generating 3D shadow art sculptures represented using both voxels and mesh. For every result in Figure \ref{fig:res} (a)-(d), we show the learned 3D sculptures (voxel and mesh based) along with the respective shadows casted from different specified directions. We could not include the optimized 3D sculpture from \cite{mitra2009shadow} as the associated object file was not downloadable through their optimization tool. We have been able to incorporate both orthogonal (Figure \ref{fig:res} (a, b, c)) and non-orthogonal views (Figure \ref{fig:res} (d) and Figure \ref{fig:teaser} (b)) to obtain the shadows that are consistent with the desired target shadow images. For a quantitative comparison, we also report IoU and Dice score. As depicted in Figure \ref{fig:res}, the IoU and Dice Score are comparable for both voxel and mesh based renderings. However, the corresponding voxel based 3D sculptures are not that smooth (low normal consistency value) when compared to those of mesh based 3D sculptures. It is important to note that the underlying voxel representation has been converted to a mesh representation to compute normal consistency values. 
While \cite{mitra2009shadow} have focused only on foreground inconsistencies (marked in orange color), we also show the background inconsistencies (marked in blue color) that appear in some of rendered shadow images. Ours is an end-to-end optimization approach without any additional editing tool to prune the generated 3D sculpture. In some cases, the mesh based approach is found to produce certain discontinuities near non-convex regions (Figure \ref{fig:res} (b,d)) for atleast one view. This is mainly attributed to the inability of icosphere to handle sharp discontinuities in the desired shape, especially when regularisation has been imposed (Equation \ref{eq:overall_loss}). The voxel based approaches may contain a few outliers (voxels outside the desired 3D shape, as marked in blue in Figure \ref{fig:res} (d)) which is generally not the case with mesh based approaches. However, the mesh based differentiable rendering method lags in handling sharp discontinuities and holes present in the shadow images. While these shortcomings are handled effectively by voxel based methods, they tend to generate discretized 3D sculpture and are often associated with high memory and computational requirements. Overall, the differentiable rendering based optimization for both the approaches has been able to generate plausible 3D shadow art sculptures and is observed to have outperformed \cite{mitra2009shadow} in handling shadow inconsistencies by a large extent without having to explicitly deform the desired shadow images.
\begin{figure*}[ht]
    \centering
    \includegraphics[width=\linewidth]{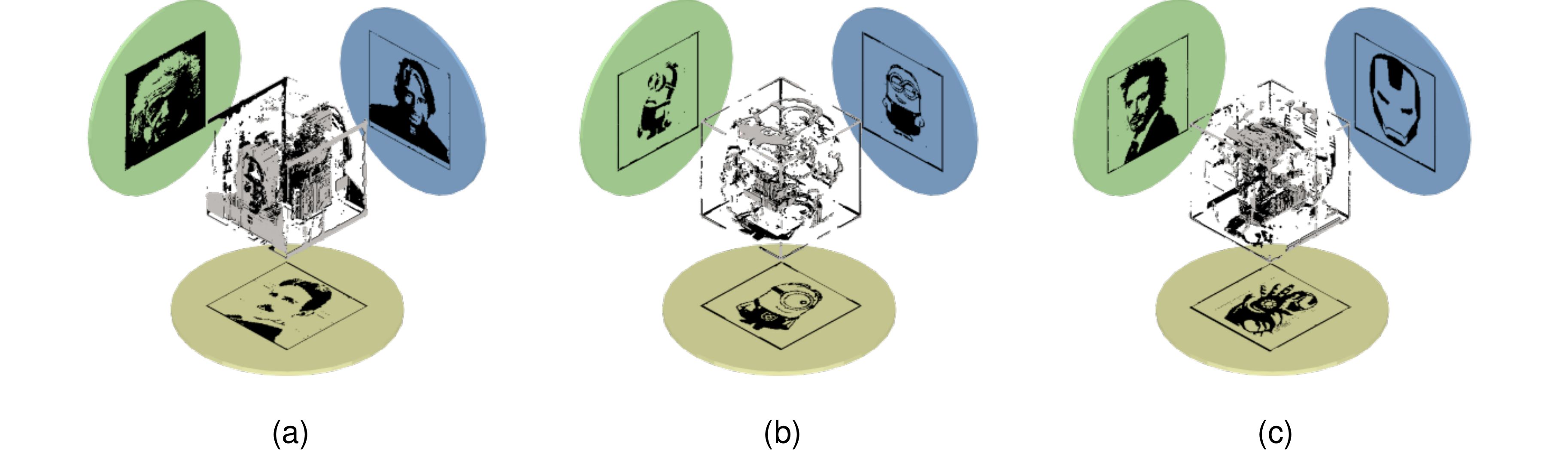}
    \caption{A seemingly random voxel soup creates three distinct shadow images of (a) \textit{Albert Einstein}, \textit{Nikola Tesla}, and \textit{APJ Abdul Kalam}, (b) \textit{Minions}, and (c) \textit{Ironman}.}
    \label{fig:faces}
\end{figure*}

\begin{figure*}[h]
    \centering
    \includegraphics[width=\linewidth]{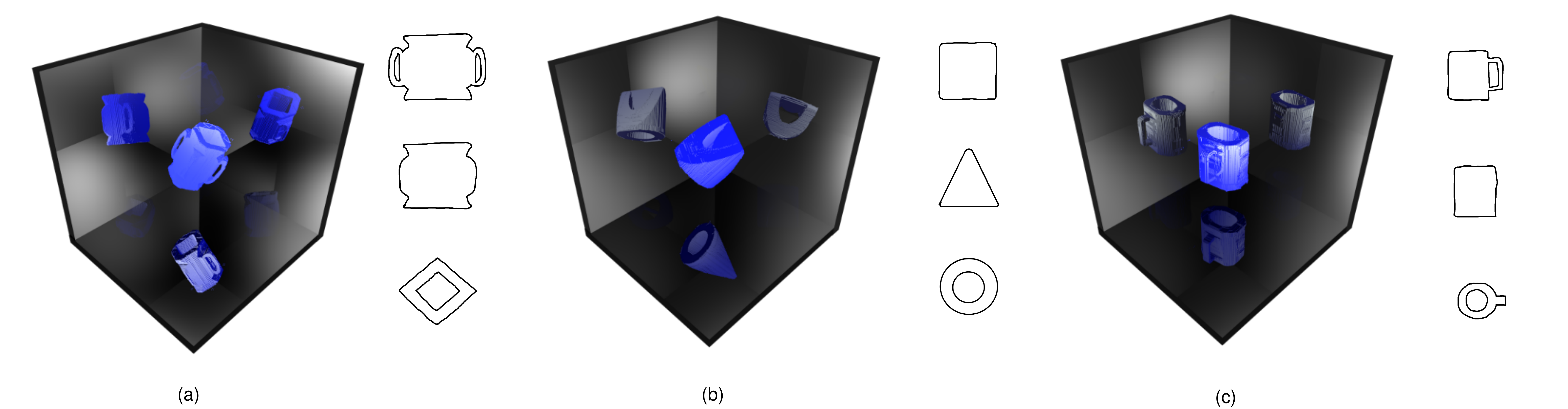}
    \caption{3D reconstruction of (a) flower vase, (b) pen-stand, and (c) coffee mug using the associated hand drawn sketches from three different views.}
    \label{fig:3D_recon}
\end{figure*}
\subsection{Comparison with the State-of-the-art method} We show the qualitative comparison of the results obtained using our voxel based differentiable rendering pipeline and the voxel based optimization tool presented in \cite{mitra2009shadow} without any deformation to the target shadow image. In Figure \ref{fig:comparison}, we observe that the shadows rendered using the proposed pipeline are highly consistent with that of the desired target shadows when compared to those produced by \cite{mitra2009shadow}. The authors of \cite{mitra2009shadow} argue that finding a consistent configuration for a given choice of input images might be impossible and hence, propose to introduce deformation in the input image so as to achieve consistency of the rendered shadow images with the desired ones. However, the differentiable rendering based optimization can handle inconsistencies without causing any explicit change in the target shadow images.

\section{Applications}\label{sec:applications}
In this section, we show some additional artistic shadow art creations and an extension to yet another application that can also benefit from our optimization approach. Figure \ref{fig:faces} depicts the creation of faces of well known scientists around the world and movie characters like \textit{Minions} and \textit{Ironman}, demonstrating the strength of differentiable rendering based optimization approach to handle complex objects or scenes with consistency. In addition to the binary silhouette images, half-toned images can also be used to generate 3D shadow art sculptures, as shown in Figure \ref{fig:teaser}. Another interesting extension is towards sketch-based modeling \cite{olsen2009sketch} where we use hand-drawn sketches of a shape from different viewpoints to automatically create the underlying 3D object. We demonstrate the creation of a flower vase (Figure \ref{fig:3D_recon} (a)), pen-stand (Figure \ref{fig:3D_recon} (b)), and a coffee mug (Figure \ref{fig:teaser} (c)) solely from hand-drawn sketches from three different views.   

\section{Conclusion}\label{sec:conclusion}
We have introduced an optimization framework for generating 3D shadow art sculptures from a set of shadow images and the associated projection information. The key idea is to explore the strength of differentiable rendering in creating visually plausible and consistent shadows of rigid objects, faces, and animated movie characters by generating the associated 3D sculpture. We have discussed both voxel and mesh-based rendering pipelines and have identified the benefits of each of them for the task at hand. Additionally, we have demonstrated the applicability of the proposed framework in reconstructing 3D shapes using their sketches drawn from three different viewpoints. At present, we have primarily considered the shadows that are associated with static sculptures and hence, themselves are static in nature. Dynamic shadow art can also be explored in near future.

\newpage

{\small
\bibliographystyle{ieee_fullname}
\bibliography{egbib}
}

\end{document}